\documentclass[sigconf,pbalance]{acmart}

\usepackage{graphicx}
\usepackage{amsmath}
\usepackage{booktabs}
\usepackage{multirow}
\usepackage{algorithm}
\usepackage{algpseudocode}
\usepackage{needspace}
\newtheorem{proposition}{Proposition}
\newtheorem{corollary}{Corollary}
\newtheorem{hypothesis}{Hypothesis}

\copyrightyear{2026}
\acmYear{2026}
\setcopyright{cc}
\setcctype{by}
\acmConference[ICVGIP 2026]{Indian Conference on Computer Vision, Graphics, and Image Processing}
{December 21--24, 2026}{Kolkata, India}
\acmBooktitle{Indian Conference on Computer Vision, Graphics, and Image Processing (ICVGIP 2026),
December 21--24, 2026, Kolkata, India}
\acmPrice{}
\acmDOI{}
\acmISBN{}

\begin{document}

\title[Coupled Scene Graph Anticipation]{Coupled Scene Graph Anticipation: When With-Constraint Recall Hides Structural Inconsistency}

\author{Nongmaithem Zeno}
\authornote{Both authors contributed equally to this research.}
\affiliation{
  \department{School of Engineering}
  \institution{Jawaharlal Nehru University}
  \city{New Delhi}
  \country{India}}
\email{nongma45\_soe@jnu.ac.in}
\orcid{0009-0003-1664-955X}

\author{Anushttha Srivastava}
\affiliation{
  \department{School of Engineering}
  \institution{Jawaharlal Nehru University}
  \city{New Delhi}
  \country{India}}
\email{anusht71\_soe@jnu.ac.in}
\orcid{0009-0000-7948-202X}

\author{Anshit Sinha}
\affiliation{
  \department{School of Engineering}
  \institution{Jawaharlal Nehru University}
  \city{New Delhi}
  \country{India}}
\email{anshit81\_soe@jnu.ac.in}
\orcid{0009-0009-7617-4203}

\author{Apoorv Sisodia}
\affiliation{
  \department{School of Engineering}
  \institution{Jawaharlal Nehru University}
  \city{New Delhi}
  \country{India}}
\email{apoorv14\_soe@jnu.ac.in}
\orcid{0009-0008-8353-4548}

\author{Suman Chowdhury}
\authornotemark[1]
\affiliation{
  \institution{Academy of Scientific and Innovative Research}
  \city{Ghaziabad}
  \state{Uttar Pradesh}
  \country{India}}
\additionalaffiliation{\institution{AI \& IoT Lab, IT Group, CSIR-Central Mechanical Engineering Research Institute}
  \city{Durgapur}
  \state{West Bengal}
  \country{India}}
\additionalaffiliation{\institution{School of Computing Technologies, RMIT University}
  \city{Melbourne}
  \state{Victoria}
  \country{Australia}}
\email{suman.cmeri21a@acsir.res.in}
\orcid{0000-0002-2166-9024}

\author{Prerana Mukherjee}
\affiliation{
  \department{School of Engineering}
  \institution{Jawaharlal Nehru University}
  \city{New Delhi}
  \country{India}}
\email{prerana@jnu.ac.in}
\orcid{0000-0003-1804-7360}

\author{Rajesh P. Barnwal}
\authornote{Corresponding author.}
\affiliation{
  \department{AI \& IoT Lab, IT Group}
  \institution{CSIR-Central Mechanical Engineering Research Institute}
  \city{Durgapur}
  \state{West Bengal}
  \country{India}}
\email{r\_barnwal@cmeri.res.in}
\orcid{0000-0002-0427-5533}
\additionalaffiliation{\institution{Academy of Scientific and Innovative Research}
  \city{Ghaziabad}
  \state{Uttar Pradesh}
  \country{India}}

\renewcommand{\shortauthors}{Nongmaithem et al.}

\begin{abstract}
Scene Graph Anticipation (SGA) predicts relationships between objects in future
video frames. Continuous-time SGA models typically represent each subject--object
pair using an independent Neural Ordinary Differential Equation (Neural ODE).
Consequently, relationships that share an entity cannot influence one another
during temporal integration, and nothing prevents mutually exclusive predictions,
such as \emph{sitting on} and \emph{standing on}, for the same subject--object
pair. To address these limitations, we introduce CoupledSGA, which connects
relationships sharing an entity through a Relationship Interaction Graph (RIG),
incorporates graph attention into the ODE vector field, and adds a Structural
Consistency Loss ($\mathcal{L}_{\mathrm{SC}}$) based on a matrix of mutually exclusive predicates.

Developing CoupledSGA revealed a fundamental limitation in the evaluation
protocol. Under the With-Constraint protocol, the evaluator retains only one
predicate from each predicate group for every subject--object pair. Consequently,
the evaluated output cannot contain both predicates involved in an exclusion rule
when those predicates belong to the same group. Since all exclusion rules
considered in our study satisfy this condition, With-Constraint Recall@$K$
(R@$K$) and mean Recall@$K$ (mR@$K$) cannot capture the violations targeted by
$\mathcal{L}_{\mathrm{SC}}$. We formally establish that these metrics remain unchanged when the
winning predicate and its score within each group are held fixed, regardless of
whether a structural violation exists.

We further derive the gradient of $\mathcal{L}_{\mathrm{SC}}$ and show that it depends on the
scores of conflicting predicates rather than their ground-truth labels. For a
two-predicate exclusion rule, the loss amplifies the ordering already established
by the model. When all scores in a group are equal, predicates involved in more exclusion rules
are suppressed more strongly, irrespective of their correctness. These properties
explain why the penalty may reduce recall even as it improves structural
consistency.

Experiments on Action Genome support this analysis. Under a matched training budget, CoupledSGA achieves 40.48\% R@20, compared with 44.78\% for
SceneSayerODE. In a controlled comparison, activating $\mathcal{L}_{\mathrm{SC}}$ after two
warm-up epochs reduces R@20 from 42.21\% to 35.18\%, while lowering the
contradiction rate under the No-Constraint protocol from 45.92\% to 4.07\%. These
results are not intended to establish state-of-the-art performance. Rather,
CoupledSGA serves as a testbed for exposing the limitations of the evaluation
protocol. Our findings demonstrate that With-Constraint recall captures the cost of
enforcing structural consistency but fails to reflect the corresponding
improvement. We therefore recommend reporting No-Constraint results alongside a
contradiction rate when evaluating methods that explicitly constrain predicate
co-activation.
\end{abstract}



\maketitle

\begin{figure*}[t]
  \centering
  \includegraphics[width=0.98\textwidth]{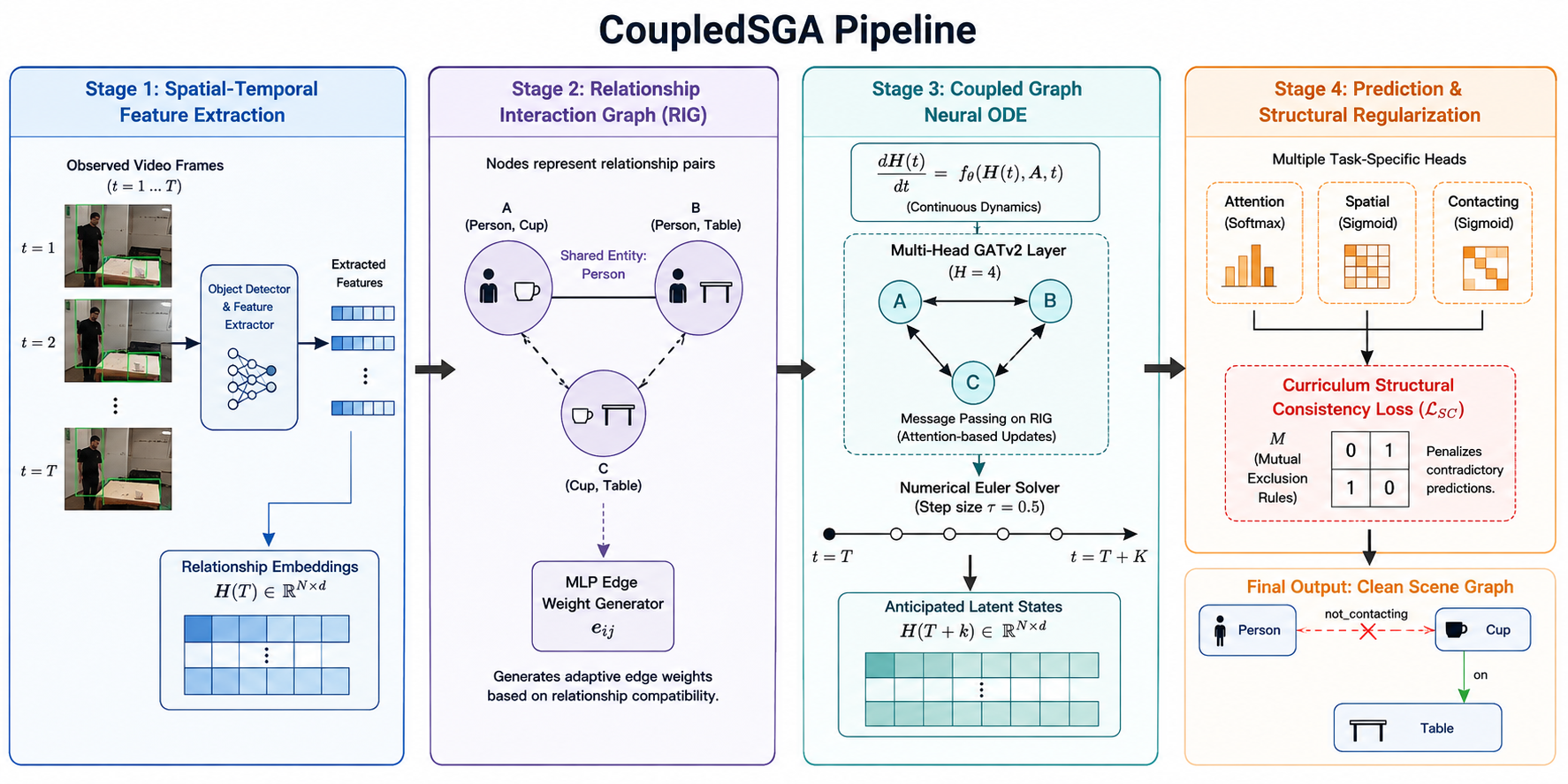}
  \Description{Four-stage pipeline: spatial-temporal feature extraction from observed
  frames, the Relationship Interaction Graph over subject-object pairs, the coupled
  graph Neural ODE with multi-head GATv2 message passing and Euler integration, and
  prediction heads regularised by the Structural Consistency Loss.}
  \caption{CoupledSGA. Observed frames give one embedding per subject--object
  relationship. The Relationship Interaction Graph connects relationships that
  share an entity. Graph attention inside the ODE vector field couples their
  trajectories during integration. Decoded predicates are regularised by the
  Structural Consistency Loss $\mathcal{L}_{\mathrm{SC}}$.}
  \label{fig:pipeline}
\end{figure*}

\section{Introduction}
\label{sec:intro}

Scene Graph Anticipation (SGA)~\cite{peddi2024scenesayer} aims to predict the
future relationships between entities in a video, representing them as
$\langle$subject, predicate, object$\rangle$ triplets. Its evaluation protocols
inherit established metrics from Scene Graph Generation
(SGG)~\cite{xu2017scene,zellers2018neural} and Video Scene Graph Generation
(VidSGG)~\cite{cong2021sttran,feng2023dsg}. In particular, Recall@$K$ (R@$K$)
reduces the penalty associated with correct predictions that are not annotated in
the ground truth~\cite{xu2017scene}, while mean Recall@$K$ (mR@$K$) mitigates
the dominance of frequent predicate classes~\cite{chen2019kern}. These metrics
are commonly reported under two evaluation protocols: With Constraint, which
restricts the number of predicates that each subject--object pair can contribute
to the evaluation, and No Constraint, which permits multiple predicates per
pair~\cite{peddi2025impartail,peddi2024scenesayer}. Although these protocols
provide complementary views of predictive performance, their ability to measure
structural consistency has received less attention.

Structural consistency is important in applications that require reliable
predictions before an interaction occurs~\cite{furnari2019rolling,gammulle2019predicting}.
For example, a person cannot simultaneously be \emph{sitting on} and
\emph{standing on} the same chair in the same relational context. Nevertheless, a
scene graph anticipation model may assign high scores to both predicates for the
same subject--object pair. For the Spatial and Contacting groups of Action
Genome~\cite{ji2020actiongenome}, predicates are decoded using independent
sigmoid outputs. Consequently, mutually exclusive predicates are not necessarily
forced to compete, and the decoder can produce structurally inconsistent
predictions. Whether such inconsistencies affect the reported evaluation results
depends on the predictions retained by the evaluation protocol.

In the evaluation implementation released with SceneSayer~\cite{peddi2024scenesayer},
the With-Constraint protocol retains the highest-scoring predicate from each
predicate group for a given subject--object pair, whereas the No-Constraint
protocol allows multiple predicates to be retained. As a result, With-Constraint
evaluation may discard evidence of a contradiction before the predictions are
scored. For example, \emph{sitting on} and \emph{standing on} both belong to the
Contacting group, so retaining only the highest-scoring predicate prevents both
from appearing in the evaluated output. Thus, a prediction may violate a structural
exclusion rule before filtering, yet the violation may remain invisible to the
reported recall metrics after filtering.

This issue emerged during the development of CoupledSGA, illustrated in
Figure~\ref{fig:pipeline}. CoupledSGA extends SceneSayerODE, the Neural Ordinary
Differential Equation (Neural ODE) variant of SceneSayer~\cite{peddi2024scenesayer},
in which each subject--object pair is modelled by an independent Neural
ODE~\cite{chen2018neural}. Because the trajectories are integrated independently,
relationships that share an entity cannot directly influence one another during
temporal evolution. CoupledSGA introduces two complementary mechanisms. First, a
Relationship Interaction Graph (RIG) connects relationships that share an entity,
and graph attention~\cite{brody2022gatv2} within the ODE vector field enables
their representations to interact during integration. Second, a Structural
Consistency Loss ($\mathcal{L}_{\mathrm{SC}}$) penalises the simultaneous activation of predicate
pairs specified as mutually exclusive in an exclusion matrix. The RIG facilitates
interaction between related relationships, whereas $\mathcal{L}_{\mathrm{SC}}$ explicitly
regularises predicate co-activation.

The central finding of this work concerns the limitations of the evaluation
protocol rather than a claim of improved predictive performance. Every exclusion
rule considered in our matrix relates two predicates within the same predicate
group. Consequently, the With-Constraint protocol cannot retain both predicates
involved in any of these violations for the same subject--object pair. We formally
prove that, when the winning predicate and its score in each group remain fixed,
With-Constraint R@$K$ and mR@$K$ are unchanged regardless of whether a
within-group exclusion rule is violated (Proposition~\ref{prop:invisible}).
Therefore, these metrics alone cannot quantify the structural improvements
targeted by $\mathcal{L}_{\mathrm{SC}}$. This result has an important implication: improvements in
structural consistency cannot be established solely through With-Constraint
recall.

We further analyse the gradient of $\mathcal{L}_{\mathrm{SC}}$ with respect to predicate scores
and show that the penalty depends on the scores of mutually exclusive predicates
rather than their ground-truth labels (Proposition~\ref{prop:gradient}). For a
two-predicate exclusion rule, the loss reinforces the ordering that the model
already assigns to the two predicates. When all scores in a group are equal, predicates involved in a greater number of exclusion rules are suppressed more strongly, irrespective
of whether they are correct. These properties explain why enforcing structural
consistency may reduce recall, particularly when the penalty is introduced before
the model has learned reliable predicate rankings.

Experiments on Action Genome are consistent with this analysis. Under a matched training budget, CoupledSGA achieves an R@20 of 40.48\%, compared with
44.78\% for SceneSayerODE. In a controlled comparison, activating $\mathcal{L}_{\mathrm{SC}}$
after two warm-up epochs reduces R@20 from 42.21\% to 35.18\%. Under the
No-Constraint protocol, the same change reduces the contradiction rate from
45.92\% to 4.07\%, while coupling alone reduces it from 54.40\% to 45.92\%. These
results are not intended to establish state-of-the-art performance. Rather, they
illustrate the trade-off between recall and structural consistency and show that
With-Constraint recall captures the cost of enforcing consistency without
reflecting the corresponding structural improvement.

The findings motivate a more informative evaluation practice for SGA methods that
explicitly regularise predicate co-activation. In addition to standard
With-Constraint metrics, evaluations should report No-Constraint results and a
contradiction rate, defined as the proportion of (subject--object pair, predicate
group) instances in which the two highest-scoring predicates form an excluded pair.
Reporting these complementary measures would help distinguish predictive recall
from structural validity and provide a clearer assessment of the effects of
consistency constraints.

\needspace{4\baselineskip}
The contributions of this work are summarised as follows:
\begin{enumerate}
  \item \textbf{An analysis of evaluation limitations.} We prove that
  With-Constraint R@$K$ and mR@$K$ cannot detect within-group structural
  violations when the winning predicate and its score in each group remain
  unchanged (Section~\ref{sec:analysis}).
  \item \textbf{A coupled SGA framework.} We introduce CoupledSGA, which uses a
  Relationship Interaction Graph and graph attention within a Neural ODE vector
  field to couple relationships that share entities, together with a Structural
  Consistency Loss to penalise mutually exclusive predicate co-activation
  (Section~\ref{sec:method}).
  \item \textbf{An analysis of the consistency penalty.} We derive the gradient of
  $\mathcal{L}_{\mathrm{SC}}$ and establish how its label-independent behaviour can reinforce
  existing predicate-score orderings in two-predicate rules and, when all scores in a
  group are equal,
  disproportionately suppress predicates involved in multiple exclusion rules
  (Section~\ref{sec:analysis}).
  \item \textbf{An empirical evaluation and reporting recommendation.} Experiments
  on Action Genome demonstrate the trade-off between recall and structural
  consistency, motivating the joint reporting of recall metrics and contradiction
  rates (Section~\ref{sec:experiments}).
\end{enumerate}

\section{Related Work}
\label{sec:related}

\paragraph{Video scene graphs and anticipation}
Static SGG extracted relationships from single images using message
passing~\cite{xu2017scene} and frequency priors~\cite{zellers2018neural}.
STTran~\cite{cong2021sttran} added factorised spatial and temporal attention for
video, and DSG-DETR~\cite{feng2023dsg} modelled long-term dependencies over object tracklets.
These models label frames they have observed, so their numbers are not a target
for anticipation. SceneSayer~\cite{peddi2024scenesayer} introduced SGA and solved
it with per-pair Neural ODEs and Neural SDEs. Action
anticipation~\cite{furnari2019rolling,gammulle2019predicting} predicts coarse
labels rather than relational structure.

\paragraph{Recent SGA}
ImparTail~\cite{peddi2025impartail} applies
curriculum-driven loss masking to VidSGG and SGA and reports SGA results under
both With-Constraint and No-Constraint settings.
FDSG~\cite{yang2025fdsg} forecasts entity labels, boxes and relationships jointly.
FORESCENE~\cite{alliegro2025forescene} predicts full future graphs with a latent
diffusion model. The last two extend SGA beyond the fixed object set and the
predicate-only prediction we study here, and none of the three is a direct
numerical baseline for our matched PredCLS budget. We note them because our evaluation
argument applies to any of them that adds a co-activation constraint.

\paragraph{Coupled dynamics on graphs}
Graph Neural ODEs~\cite{poli2019graph}, CG-ODE~\cite{huang2021cgode} and
Graph-Coupled Oscillator Networks~\cite{rusch2022graph} model interacting units in
continuous time. They couple over the entity graph and predict a continuous state.
We couple over a graph of relationships, which is where mutual exclusion is
defined, and our target is a discrete symbolic output.

\paragraph{Structural constraints and scheduled objectives}
Symbolic structure has been added to visual predictors through knowledge
graphs~\cite{zareian2020bridging}, probabilistic
logic~\cite{manhaeve2018deepproblog} and energy-based penalties on invalid
subgraphs~\cite{suhail2021energy}. TEMPURA~\cite{nag2023tempura} addresses
long-tailed predicates with memory-guided training. Our schedule borrows the idea of curriculum
learning~\cite{bengio2009curriculum}, which presents easier training examples
before harder ones, and applies it to an auxiliary objective;
ImparTail~\cite{peddi2025impartail} uses a curriculum to mask head-class predicate
losses. Our penalty does not act on the class
distribution, and it is narrower than the symbolic approaches: it constrains only
which predicates may be simultaneously active on one pair.

\section{Task and Evaluation Protocol}
\label{sec:task}

\subsection{Scene Graph Anticipation}
\label{sec:sga}

Let $I_1, \dots, I_T$ be observed frames. SGA predicts the relationships holding in
$K$ later frames $I_{T+1}, \dots, I_{T+K}$. A \emph{relationship pair} is an
ordered pair of entities $(s, o)$; a \emph{predicate} is a relation class assigned
to that pair. We write $P_T$ for the set of pairs active at the last observed frame
and $N = |P_T|$.

We use Predicate Classification (PredCLS): entity identities and boxes are given
for all frames, and only the predicate must be predicted. The released Action
Genome annotations~\cite{ji2020actiongenome} define 26 predicates in three
\emph{predicate groups}: Attention (3 classes, decoded with a softmax and so single-label by
construction), Spatial (6 classes, multi-label) and Contacting (17 classes,
multi-label). Spatial and Contacting are decoded with independent sigmoids, so
nothing forces the classes inside a group to compete.

\subsection{The With-Constraint Emission Rule}
\label{sec:emission}

Our main result depends on the exact scoring rule, so we state it before anything
else. Under \emph{With Constraint}, for each pair $i$ and each predicate group $g$,
the evaluator keeps only the top-scoring predicate of that group:
\begin{equation}
\label{eq:emission}
\hat{p}_i^{\,g} = \arg\max_{u \in g} \ \hat{y}_{i,u},
\end{equation}
where $\hat{y}_{i,u}$ is the predicted score of predicate $u$ for pair $i$. The
kept triplets are ranked by score, and R@$K$ is the fraction of ground-truth
triplets in the top $K$. Under \emph{No Constraint}, several predicates per group
are kept, so co-activated predicates can both appear. We use the emission rule
implemented in the Action Genome evaluation code released with prior SGA
work~\cite{peddi2024scenesayer}; the results in Section~\ref{sec:analysis} concern
the rule itself and not any historical label attached to it.

\subsection{Two Notions of Consistency}
\label{sec:notions}

Let $M \in \{0,1\}^{C \times C}$ be a symmetric, zero-diagonal \emph{exclusion
matrix} over the $C$ Spatial and Contacting classes, with $M_{uv} = 1$ when $u$
and $v$ cannot both hold of one pair. Call $(u, v)$ an \emph{excluded pair}, and
call the rule \emph{within-group} when $u$ and $v$ belong to the same predicate
group. We keep two notions separate throughout, because conflating them is easy
and misleading.

\textbf{Structural validity} is a property of a decoded graph: no pair is assigned
both members of any excluded pair. It is evaluated after emission.

\textbf{Co-activation mass} is a property of the continuous scores, before
emission:
\begin{equation}
\label{eq:cam}
c_i = \sum_{u<v} M_{uv}\, \hat{y}_{i,u}\, \hat{y}_{i,v} \;\geq\; 0.
\end{equation}
It is zero exactly when, for every excluded pair, at least one of the two scores is
zero.

The two do not coincide. A decoded graph can be structurally valid while
$c_i > 0$, and a small $c_i$ does not certify validity. Our loss minimises
co-activation mass; what a downstream consumer cares about is validity. Keeping
the distinction visible is what makes the analysis in Section~\ref{sec:analysis}
precise.

\section{CoupledSGA}
\label{sec:method}

\subsection{Relation Encoder}
\label{sec:encoder}

For each observed frame $t$ and pair $(s, o)$, a spatial-temporal
backbone~\cite{cong2021sttran} pools RoIAlign features for the subject box, the
object box and their union, concatenates a relative-box encoding and learned class
embeddings, and refines the result with spatial attention across pairs and
temporal attention across frames. This yields $h_{so}^{(t)} \in \mathbb{R}^{d}$
with $d = 1936$. Stacking the pairs active at frame $T$ gives the initial state
$H(T) \in \mathbb{R}^{N \times d}$ for integration.

\subsection{Relationship Interaction Graph}
\label{sec:rig}

The RIG is a graph whose nodes are the $N$ relationship pairs. Two nodes are joined
when the underlying pairs share an entity:
\begin{equation}
\label{eq:adjacency}
A_{ij} =
\begin{cases}
1, & \text{if } s_i = s_j \,\lor\, s_i = o_j \,\lor\, o_i = s_j \,\lor\, o_i = o_j, \\
0, & \text{otherwise},
\end{cases}
\end{equation}
with self-loops added, $\tilde{A} = A + I_N$. We also learn a scalar weight per edge
from the two incident embeddings,
\begin{equation}
\label{eq:edge}
e_{ij} = \mathrm{MLP}_{\mathrm{edge}}\big(\, [\, h_i(T) \,\|\, h_j(T) \,] \,\big),
\end{equation}
which enters the attention score in Equation~\ref{eq:attention}. Action Genome
annotates person--object relations, so most pairs in a frame share the same actor
and the RIG is close to complete. Topology alone therefore carries little
information on this dataset. Which neighbours matter is decided by $e_{ij}$ and by
the attention weights, not by the adjacency. On multi-agent data, the same rule
would produce a block-structured graph.

\subsection{Coupled Graph Neural ODE}
\label{sec:ode}

We integrate the relationship states from $T$ to the prediction horizon by solving
$dH(t)/dt = f_\theta(H(t), \tilde{A}, t)$ with $H(T)$ as the initial condition. The
point of CoupledSGA is that $f_\theta$ does not act row by row. At every solver
step, each relationship updates its state from its RIG neighbours:
\begin{equation}
\label{eq:ode}
\frac{d h_i(t)}{dt} = \sigma\!\left( \sum_{j \in \mathcal{N}(i)} \alpha_{ij}(t)\, W_v\, h_j(t) \right) - \gamma\, h_i(t),
\end{equation}
where $\mathcal{N}(i)$ are the neighbours of $i$ in $\tilde{A}$, $W_v$ is a shared
value projection, $\sigma$ is SiLU, and $-\gamma h_i(t)$ with $\gamma = 1$ damps the
state, which helps keep trajectories bounded over the rollout. The attention weights are
GATv2~\cite{brody2022gatv2} scores, extended with the learned edge weight and a
time embedding:
\begin{multline}
\label{eq:attention}
\alpha_{ij}(t) = \operatorname{softmax}_{j \in \mathcal{N}(i)} \Bigl( \mathrm{LeakyReLU}\bigl( a^{\top} [\, W_q h_i(t) \,\|\, W_k h_j(t) \\
\,\|\, \phi(t) \,] + w_e\, e_{ij} \bigr) \Bigr),
\end{multline}
with $\phi(t)$ a Fourier embedding of the integration time and $w_e$ a scalar
parameter. Including $\phi(t)$ lets the coupling strength change along the horizon
rather than staying fixed at its value at $t = T$. We use two layers and four
heads, averaged to preserve $d$.

We solve the system with explicit Euler at a fixed step $\tau = 0.5$, so the number
of derivative evaluations is $S = K/\tau$, set by the horizon and not by the scene.

\paragraph{Cost}
Each solver step costs $\mathcal{O}(N d^2)$ for the projections and
$\mathcal{O}(|\mathcal{E}_{\mathrm{RIG}}|\, d)$ for message passing, against
$\mathcal{O}(N d^2)$ per step for an uncoupled model. The marginal cost of coupling
is the second term. With $d = 1936$ and few active pairs per frame, it is dominated
by the projections. This is an analytical statement; we do not report measured
runtime.

\subsection{Structural Consistency Loss}
\label{sec:loss}

\begin{table}[t]
\centering
\caption{Exclusion rules in $M$. \emph{Deg.} is the number of rules the predicates
of that row appear in. Every rule is within-group, which
Proposition~\ref{prop:invisible} uses; the uneven degrees are what
Corollary~\ref{cor:degree} uses.}
\label{tab:rules}
\small
\begin{tabular}{@{}llc@{}}
\toprule
\textbf{Group} & \textbf{Excluded pair} & \textbf{Deg.} \\
\midrule
Spatial    & above / beneath           & 1\,/\,1 \\
Spatial    & in front of / behind      & 1\,/\,1 \\
Contacting & sitting on / standing on  & 1\,/\,1 \\
Contacting & wiping / drinking from    & 1\,/\,1 \\
Contacting & holding / not contacting  & 1\,/\,3 \\
Contacting & touching / not contacting & 1\,/\,3 \\
Contacting & carrying / not contacting & 1\,/\,3 \\
\bottomrule
\end{tabular}
\end{table}

Let $\hat{Y} \in [0,1]^{N \times C}$ hold the sigmoid scores of the Spatial and
Contacting heads at a predicted timestep, $C = 23$. Summing the co-activation mass
of Equation~\ref{eq:cam} over pairs gives
\begin{equation}
\label{eq:lsc}
\mathcal{L}_{\mathrm{SC}} = \frac{1}{N} \operatorname{Tr}\big(\hat{Y} M \hat{Y}^{\top}\big) = \frac{2}{N} \sum_{i=1}^{N} c_i \;\geq\; 0,
\end{equation}
which is zero exactly when every excluded pair has at least one score of zero. This
is the co-activation condition of Section~\ref{sec:notions}, not discrete
structural validity. The loss adds no cost at inference. Table~\ref{tab:rules}
lists the rules in $M$. Two columns matter later: every rule is within-group, and
\emph{not contacting} appears in three rules while every other predicate appears in
one.

\paragraph{Schedule}
Section~\ref{sec:analysis} shows that $\mathcal{L}_{\mathrm{SC}}$ reinforces the
ordering the network already holds. That argues for delaying it until the ordering
is reliable, so we gate it by epoch: $\lambda_{\mathrm{SC}}^{(e)} = 0$ for
$e < e_{\mathrm{start}}$ and $\lambda_{\mathrm{SC}}$ afterwards. The full objective
is
\begin{equation}
\label{eq:total}
\mathcal{L}_{\mathrm{total}} = \mathcal{L}_{\mathrm{obs}} + \mathcal{L}_{\mathrm{ant}} + \beta_1 \mathcal{L}_{\mathrm{recon}} + \beta_2 \mathcal{L}_{\mathrm{bbox}} + \lambda_{\mathrm{SC}}^{(e)} \mathcal{L}_{\mathrm{SC}},
\end{equation}
where $\mathcal{L}_{\mathrm{obs}}$ and $\mathcal{L}_{\mathrm{ant}}$ are
cross-entropy and binary cross-entropy terms on observed and predicted predicates,
$\mathcal{L}_{\mathrm{recon}}$ is an $L_1$ latent reconstruction term and
$\mathcal{L}_{\mathrm{bbox}}$ is a smooth-$L_1$ box term ($\beta_1 = 1$,
$\beta_2 = 0.1$). Algorithm~\ref{alg:training} gives the loop.

\begin{algorithm}[t]
\caption{CoupledSGA training}
\label{alg:training}
\begin{algorithmic}[1]
\Require observed frames, entities, horizon $K$, step $\tau$, threshold $e_{\mathrm{start}}$, weight $\lambda_{\mathrm{SC}}$
\For{$e = 0$ \textbf{to} $E-1$}
    \State $\lambda_{\mathrm{SC}}^{(e)} \gets \lambda_{\mathrm{SC}}$ \textbf{if} $e \geq e_{\mathrm{start}}$ \textbf{else} $0$
    \For{each video batch}
        \State encode observed frames; form $H(T)$
        \State build $\tilde{A}$ and $e_{ij}$ \Comment{Eq.~\ref{eq:adjacency}, \ref{eq:edge}}
        \For{$k = 1$ \textbf{to} $K$}
            \State integrate Eq.~\ref{eq:ode} by Euler steps of size $\tau$ to $T+k$
            \State decode $\hat{Y}_{T+k}$
        \EndFor
        \State update by AdamW on $\mathcal{L}_{\mathrm{total}}$ \Comment{Eq.~\ref{eq:total}}
    \EndFor
\EndFor
\end{algorithmic}
\end{algorithm}

\section{Why With-Constraint Recall Misses Structural Violations}
\label{sec:analysis}

This section contains the paper's two analytical results. Both follow from
definitions already given, and neither depends on our experiments.

\subsection{The Evaluator Discards the Evidence}
\label{sec:invisible}

A contradiction needs two predicates to be present at once. With Constraint keeps
one predicate per group, so the evaluator removes the evidence of the
contradiction before recall is computed.

\begin{proposition}[Within-group violations are invisible]
\label{prop:invisible}
Let $(u, v)$ be an excluded pair with $u, v$ in the same predicate group $g$. For
any score matrix $\hat{Y}$, the With-Constraint output of
Equation~\ref{eq:emission} contains at most one of $u$ and $v$ for any pair $i$.
Consequently, R@$K$ and mR@$K$ computed under With Constraint depend on the scores of group $g$ only through its $\arg\max$ and the score of that $\arg\max$: changing the score of the other member of an excluded pair, with the $\arg\max$ and its score held fixed, leaves both metrics unchanged.
\end{proposition}

\begin{proof}
Equation~\ref{eq:emission} keeps a single element of $g$ per pair, namely
$\arg\max_{w \in g} \hat{y}_{i,w}$. Since $u, v \in g$, at most one of them is
kept. The scored candidate set therefore never contains both, for any values of $\hat{y}_{i,u}$ and $\hat{y}_{i,v}$. It holds only the $\arg\max$ of each group and its score, so a statistic computed from that set cannot depend on the score of a predicate that is not kept.
\end{proof}

Every rule in Table~\ref{tab:rules} is within-group. The quantity
$\mathcal{L}_{\mathrm{SC}}$ minimises is thus one that With-Constraint recall does
not respond to. The violations are nonetheless real and do appear under No
Constraint, where more than one predicate per group is kept. A study that claims
to improve predicate co-activation consistency therefore cannot support that
claim with With-Constraint recall alone.

\subsection{The Penalty Does Not Know Which Predicate Is Correct}
\label{sec:gradient}

\begin{proposition}[Gradient of the exclusion penalty]
\label{prop:gradient}
Define the conflict mass of predicate $u$ at pair $i$ as
$m_{i,u} = \sum_{v} M_{uv}\, \hat{y}_{i,v}$. Then
\begin{equation}
\label{eq:grad}
\frac{\partial \mathcal{L}_{\mathrm{SC}}}{\partial \hat{y}_{i,u}} = \frac{2}{N}\, m_{i,u}.
\end{equation}
Writing $\hat{y}_{i,u} = \sigma(z_{i,u})$ for the head logits and
$g_{i,u} = \hat{y}_{i,u}(1 - \hat{y}_{i,u})$, a gradient step of size $\eta$ on $\mathcal{L}_{\mathrm{SC}}$
changes the logit gap of an excluded pair $(u, v)$ by
\begin{equation}
\label{eq:gap}
\Delta\big(z_{i,u} - z_{i,v}\big) = \frac{2\eta}{N}\big(m_{i,v}\, g_{i,v} - m_{i,u}\, g_{i,u}\big).
\end{equation}
No term depends on the ground-truth label.
\end{proposition}

\begin{proof}
$M$ is symmetric with zero diagonal, so differentiating Equation~\ref{eq:lsc} gives
Equation~\ref{eq:grad}. The chain rule through the sigmoid gives
\[
\frac{\partial \mathcal{L}_{\mathrm{SC}}}{\partial z_{i,u}} = \frac{2}{N}\, m_{i,u}\, g_{i,u},
\qquad
\Delta z_{i,u} = -\frac{2\eta}{N}\, m_{i,u}\, g_{i,u}
\]
under gradient descent. Subtracting the two updates gives Equation~\ref{eq:gap}.
Equation~\ref{eq:lsc} contains no label term.
\end{proof}

Two consequences follow, and they cover the two kinds of rows in
Table~\ref{tab:rules}.

\begin{corollary}[Two-predicate rules: the ordering is reinforced]
\label{cor:order}
If $u$ and $v$ are each other's only excluded partner, then
$m_{i,u} = \hat{y}_{i,v}$ and $m_{i,v} = \hat{y}_{i,u}$, and
Equation~\ref{eq:gap} reduces to
\begin{equation}
\label{eq:order}
\Delta\big(z_{i,u} - z_{i,v}\big) = \frac{2\eta}{N}\, \hat{y}_{i,u}\, \hat{y}_{i,v}\, \big(\hat{y}_{i,u} - \hat{y}_{i,v}\big),
\end{equation}
which has the sign of $\hat{y}_{i,u} - \hat{y}_{i,v}$. The penalty widens whichever
gap already exists.
\end{corollary}

Concretely, if the model scores \emph{sitting on} at 0.8 and \emph{standing on} at
0.2 for one pair, the penalty pushes the 0.2 down harder than the 0.8. It would do
exactly the same if \emph{standing on} were the correct label. The loss removes a
contradiction by amplifying whichever side is currently ahead.

\begin{corollary}[Multi-partner rules: a bias by degree]
\label{cor:degree}
Let $u$ and $v$ be excluded partners with $\hat{y}_{i,u} = \hat{y}_{i,v}$, so
$g_{i,u} = g_{i,v}$. Then Equation~\ref{eq:gap} has the sign of
$m_{i,v} - m_{i,u}$: the tie is broken toward the predicate with the smaller
conflict mass. If all scores in the group are equal, $m_{i,u}$ is proportional to
the number of rules $u$ appears in, so ties break toward predicates of lower degree
in $M$.
\end{corollary}

In our matrix, \emph{not contacting} has degree three and its three partners have
degree one (Table~\ref{tab:rules}). The penalty therefore breaks ties away
from \emph{not contacting} and toward the contact predicates,
independently of which is correct. This is a property of the quadratic form, not of
the data. It suggests that an exclusion penalty of this shape should be normalised
by degree, or replaced by a within-group softmax over excluded sets. We did not
test either variant.

\begin{hypothesis}[Expected effect on With-Constraint recall]
\label{cor:recall}
Under With Constraint, R@$K$ depends on the scores only through the within-group
$\arg\max$ and the ranking of kept triplets. By Corollaries~\ref{cor:order}
and~\ref{cor:degree}, $\mathcal{L}_{\mathrm{SC}}$ moves the $\arg\max$ for reasons
unrelated to the label. Applied while the ordering is unreliable, it consolidates
whichever errors are present; applied once the ordering is largely correct, its
effect on the $\arg\max$ is small. We therefore hypothesise that its contribution to With-Constraint recall is
non-positive early in training and near zero later; this does not follow from the
gradient alone, since recall also depends on the supervised terms.
\end{hypothesis}

We separate three things here. The \emph{intended role} of
$\mathcal{L}_{\mathrm{SC}}$ is to reduce co-activation mass. Its \emph{unintended
optimisation effect} is to reinforce the current ordering and to penalise
high-degree predicates. Hypothesis~\ref{cor:recall} is a \emph{prediction} about
recall, not a measurement: it says that enabling the penalty early should depress
With-Constraint recall relative to leaving it off. Section~\ref{sec:controlled}
measures the effect of enabling it after two warm-up epochs.

\section{Experiments}
\label{sec:experiments}

\subsection{Setup}
\label{sec:setup}

\begin{table}[t]
\centering
\caption{Experimental conditions, shared by every run.}
\label{tab:conditions}
\small
\begin{tabular}{@{}ll@{}}
\toprule
Dataset / task     & Action Genome, PredCLS \\
Split              & 600 train / 200 test videos \\
Prediction horizon & $K = 3$ frames \\
Sequence cap       & 20 frames \\
Optimiser          & AdamW~\cite{loshchilov2019decoupled} \\
Learning rate      & $1 \times 10^{-5}$, weight decay $10^{-4}$ \\
Gradient clipping  & norm 5.0 \\
Epochs             & 4 \\
Runs               & 1 (Table~\ref{tab:matched}); 3 seeds (Table~\ref{tab:controlled}) \\
ODE solver         & explicit Euler, $\tau = 0.5$ \\
Baseline           & SceneSayerODE~\cite{peddi2024scenesayer} \\
Protocol           & With Constraint; No Constraint (Table~\ref{tab:controlled}) \\
\bottomrule
\end{tabular}
\end{table}

We use Action Genome~\cite{ji2020actiongenome}, built on
Charades~\cite{sigurdsson2016hollywood}. Compute was one NVIDIA RTX~3060 with
12\,GB, which bounds what these experiments can establish. We fix an 800-video
subset and train every model on it with identical features, optimiser and
evaluation code, so the comparison is like-for-like even though the absolute
numbers are not comparable to published full-split results.
Table~\ref{tab:conditions} lists the conditions.

The 600 training and 200 test videos are drawn from the official Action Genome
train and test partitions, respectively, so no test video is seen during
training.\footnote{Code and the exact video identifiers of the subset are
available at \url{https://github.com/zenoNong/CoupledSGA}, so that the split can
be reconstructed exactly.}

We compare against SceneSayerODE, the Neural ODE variant of
SceneSayer~\cite{peddi2024scenesayer}, since it differs from CoupledSGA in
precisely the property we study: whether the vector field acts on each pair
independently. We report R@$K$, mR@$K$ averaged over the 26 predicate classes, and
their harmonic mean hR@$K$, for cutoffs $K \in \{10, 20, 50\}$ (unrelated to the horizon $K = 3$).
Proposition~\ref{prop:invisible} says that none of these responds to the
violations $\mathcal{L}_{\mathrm{SC}}$ targets. We report them because R@$K$ and mR@$K$ are the standard SGA
metrics~\cite{peddi2024scenesayer}, and hR@$K$ combines the two. Section~\ref{sec:controlled} adds a structural measurement
under No Constraint, and Section~\ref{sec:implications} returns to what should
accompany them.

\begin{figure*}[t]
  \centering
  \includegraphics[width=0.95\textwidth]{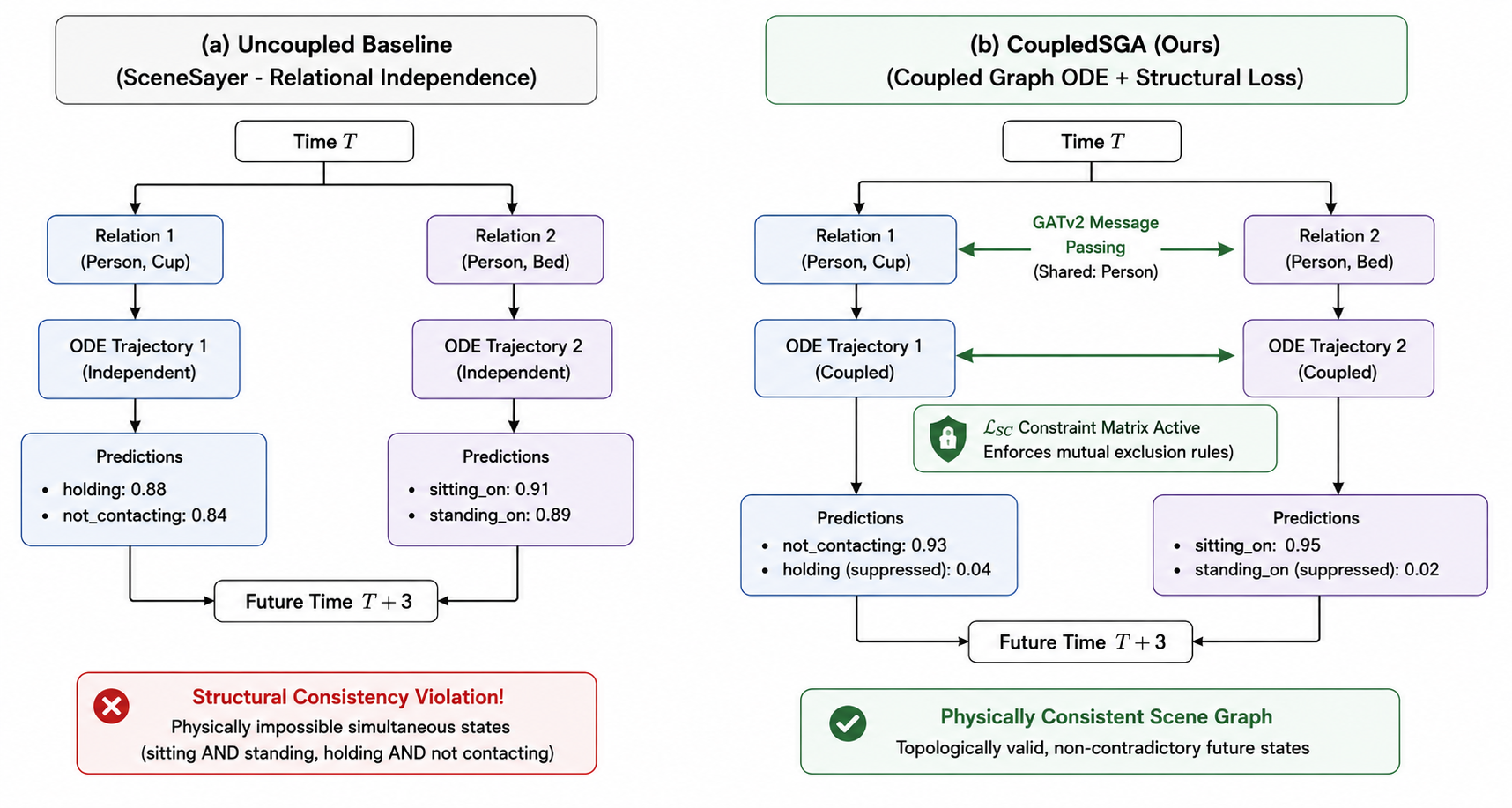}
  \Description{Two panels. Left: an uncoupled ODE assigns high scores to mutually
  exclusive predicates for the same pair. Right: CoupledSGA couples the trajectories
  over the RIG and penalises co-activation of excluded predicates.}
  \caption{Illustration of the target failure mode; the scores shown are
  illustrative and are not model outputs. Left: an uncoupled ODE evolves each relationship alone, and nothing in it prevents high scores on mutually exclusive
  predicates such as \emph{sitting on} and \emph{standing on}. Right: CoupledSGA
  couples the trajectories and penalises co-activation. By
  Proposition~\ref{prop:invisible}, this difference does not appear in
  With-Constraint recall (Table~\ref{tab:matched}); the contradiction rate of
  Table~\ref{tab:controlled} measures it.}
  \label{fig:failure}
\end{figure*}

\subsection{Matched-Budget Comparison}
\label{sec:matched}

\begin{table}[t]
\centering
\caption{Matched-budget comparison (Action Genome, PredCLS, With Constraint). Both
models use the conditions of Table~\ref{tab:conditions}. These are small-subset
results under a four-epoch budget and are not a full-benchmark comparison.}
\label{tab:matched}
\small
\begin{tabular}{@{}lcccc@{}}
\toprule
\textbf{Method} & $K$ & \textbf{R@$K$} & \textbf{mR@$K$} & \textbf{hR@$K$} \\
\midrule
\multirow{3}{*}{SceneSayerODE~\cite{peddi2024scenesayer}}
  & 10 & \textbf{43.22} & \textbf{10.79} & \textbf{17.27} \\
  & 20 & \textbf{44.78} & \textbf{11.47} & \textbf{18.26} \\
  & 50 & \textbf{44.78} & \textbf{11.48} & \textbf{18.27} \\
\midrule
\multirow{3}{*}{CoupledSGA (ours)}
  & 10 & 38.96 & 9.51  & 15.29 \\
  & 20 & 40.48 & 10.10 & 16.16 \\
  & 50 & 40.49 & 10.10 & 16.17 \\
\bottomrule
\end{tabular}
\end{table}

On our matched 600/200 split, CoupledSGA does not outperform SceneSayerODE under
With-Constraint recall. Table~\ref{tab:matched} gives the numbers. The baseline
leads by 4.30 points of R@20, 1.37 points of mR@20 and 2.10 points of hR@20, and by
similar margins at $K = 10$.

\paragraph{The gap mixes coupling and penalty}
The CoupledSGA row of Table~\ref{tab:matched} is taken after
$\mathcal{L}_{\mathrm{SC}}$ was switched off, from a run in which it was active
($\lambda_{\mathrm{SC}} = 0.1$) in epochs 0--1. The difference in
Table~\ref{tab:matched} therefore does not isolate the coupled vector field.
Section~\ref{sec:controlled} separates the two effects with controlled runs.

\paragraph{Capacity and budget are confounded}
The GATv2 field adds three $d \times d$ projections per head per layer at
$d = 1936$, many times the parameters of the row-wise field it replaces. Four
epochs at a learning rate of $10^{-5}$ is a short schedule for that many new
parameters, and the added capacity is what is least likely to be trained under it.
We cannot separate a real deficit of the coupled formulation from an undertrained
one, and we do not claim to.

\paragraph{\texorpdfstring{$K = 50$}{K = 50} is saturated}
R@20 and R@50 agree to within 0.01 points for both models, because each pair
contributes one triplet per group, and a frame with few active pairs yields a
candidate set that top-20 already exhausts. Only $K = 10$ discriminates at this
scale.

\subsection{Controlled Comparison under Both Protocols}
\label{sec:controlled}

\begin{table}[t]
\centering
\caption{Controlled comparison (Action Genome, PredCLS; conditions of
Table~\ref{tab:conditions}). R@20 is computed under With Constraint and the
contradiction rate CR (Equation~\ref{eq:cr}) under No Constraint, on anticipated
frames. Runs B and C differ only in whether $\mathcal{L}_{\mathrm{SC}}$
($\lambda_{\mathrm{SC}} = 0.1$) is active in epochs 2--3; A and B differ only in
coupling. Mean $\pm$ standard deviation over three seeds.}
\label{tab:controlled}
\small

\begin{tabular}{@{}lllcc@{}}
\toprule
 & \textbf{Model} & $\mathcal{L}_{\mathrm{SC}}$ & \textbf{R@20 (\%)} $\uparrow$ & \textbf{CR (\%)} $\downarrow$ \\
\midrule
A & SceneSayerODE & --- & \textbf{44.15}\,$\pm$\,0.51 & 54.40\,$\pm$\,1.21 \\
B & CoupledSGA    & off & 42.21\,$\pm$\,0.75 & 45.92\,$\pm$\,1.44 \\
C & CoupledSGA    & epochs 2--3 & 35.18\,$\pm$\,1.10 & \textbf{4.07}\,$\pm$\,0.46 \\
\bottomrule
\end{tabular}
\end{table}

Section~\ref{sec:matched} leaves two questions open: how much of the gap in
Table~\ref{tab:matched} is due to coupling and how much to the penalty, and whether $\mathcal{L}_{\mathrm{SC}}$ reduces contradictions at all. We answer both
with three runs trained under the conditions of Table~\ref{tab:conditions}:
SceneSayerODE (A), CoupledSGA with $\lambda_{\mathrm{SC}} = 0$ throughout (B), and
CoupledSGA with $\lambda_{\mathrm{SC}} = 0.1$ from $e_{\mathrm{start}} = 2$ (C),
which is the schedule of Section~\ref{sec:loss}. B and C share the architecture,
settings and budget, and differ only in whether $\mathcal{L}_{\mathrm{SC}}$ is
active in epochs 2--3; A and B differ only in coupling.

Each run is scored twice. R@20 is computed under With Constraint, as in
Table~\ref{tab:matched}. Contradictions are measured under No Constraint, where they can reach the output,
by the contradiction rate (CR), a rank-based proxy for structural validity. For
pair $i$ and group $g$ (Spatial or Contacting), let $u_1^{i,g}$ and $u_2^{i,g}$ be
the two highest-scoring predicates of $g$. Then
\begin{equation}
\label{eq:cr}
\mathrm{CR} = \frac{100}{|\mathcal{I}|} \sum_{(i,g) \in \mathcal{I}} M_{u_1^{i,g} u_2^{i,g}} \ (\%),
\end{equation}
where $\mathcal{I}$ collects all (pair, group) instances over the anticipated
frames. CR uses ranks only, so it needs no threshold. Table~\ref{tab:controlled}
reports the results. Its runs were trained separately from those of
Table~\ref{tab:matched}; SceneSayerODE reaches 44.15 R@20 here (mean of three seeds) and 44.78 there under identical conditions.

\paragraph{The loss reduces contradictions}
Between B and C, enabling $\mathcal{L}_{\mathrm{SC}}$ lowers CR from 45.92\% to
4.07\%, roughly elevenfold, and lowers R@20 by 7.03 points, from 42.21 to 35.18.
This is Proposition~\ref{prop:invisible} made measurable: With-Constraint recall
registers the cost of the loss but not its structural gain, and
Table~\ref{tab:controlled} reports both.

\paragraph{Coupling alone reduces contradictions}
Without any knowledge of excluded pairs, the RIG lowers CR from 54.40\% to 45.92\%
(A to B), at a cost of 1.94 points of R@20.

\paragraph{The schedule does not avoid the cost}
Hypothesis~\ref{cor:recall} predicts that the cost shrinks once the ordering within
excluded pairs is largely correct. The measured cost suggests that, at our budget, two warm-up epochs did not reach
that regime.

\subsection{Implications for Evaluation}
\label{sec:implications}

Figure~\ref{fig:failure} illustrates the failure mode the loss targets.
With-Constraint recall shows only the cost of the penalty, because it does not
respond to co-activation; the benefit appears only under No Constraint
(Table~\ref{tab:controlled}). What follows is a recommendation for any SGA study
that constrains which predicates may be simultaneously active.

Report results under No Constraint as well, since that is where co-activated
predicates reach the scored output at all. Alongside recall, report two quantities
computable from stored predictions without further training. A \emph{contradiction
rate} (Equation~\ref{eq:cr}) is the fraction of (pair, group) instances whose two
highest-scoring predicates form an excluded pair; it compares ranks, so it needs no
threshold. A \emph{structural consistency score} is the fraction of predicted
graphs containing no excluded co-activation; it is stricter and closer to what a
downstream consumer needs, since consumers read graphs rather than single edges.
We did not find either reported alongside With-Constraint recall in the SGA
studies cited here. We report the first in Table~\ref{tab:controlled} and leave the second to
future work.

\section{Discussion and Limitations}
\label{sec:limitations}

The empirical claims are bounded by scale. All results come from an 800-video
subset and a four-epoch budget; Table~\ref{tab:matched} comes from a single seed
and Table~\ref{tab:controlled} from three seeds. Propositions~\ref{prop:invisible}
and~\ref{prop:gradient} do not depend on scale. The four-epoch budget leaves the
capacity confound of Section~\ref{sec:matched} unresolved, and two warm-up epochs
were not enough for the schedule of Section~\ref{sec:loss} to avoid the recall cost
(Section~\ref{sec:controlled}). Structural validity is assessed only through the top-two contradiction rate; the structural consistency score of
Section~\ref{sec:implications} is not reported. The exclusion matrix is
hand-curated, and its coverage of the Action Genome ontology is unknown, so the
contradiction rate inherits that coverage. Explicit Euler at $\tau = 0.5$ is
first-order, and its truncation error is not quantified against an adaptive solver.

Two failure patterns recur in the predicted graphs. Under heavy occlusion, the
backbone features degrade before the RIG is built, and coupling then spreads the
resulting noise across neighbouring trajectories instead of damping it. For the
rarest contacting predicates, which have fewer than ten training instances in the
600-video split, neither model predicts transitions reliably.

Because Action Genome is built from Charades~\cite{sigurdsson2016hollywood}, whose
activity videos participants recorded in their own homes, models that forecast
human interactions could be repurposed for surveillance. We use only existing
annotations and collect no new data.

\section{Conclusion}
\label{sec:conclusion}

We set out to remove independent relation trajectories from continuous-time Scene
Graph Anticipation, and built CoupledSGA: a graph over relationships, coupled into
the vector field of a Neural ODE, with a penalty on mutually exclusive predicates.
Building it exposed a problem with the measurement rather than the model.

What we prove: under the With-Constraint emission rule, a scored output can never
contain both members of a within-group exclusion rule, so recall computed under that protocol cannot register those violations. What we derive: the gradient of a
quadratic exclusion penalty is label-free, it widens whichever ordering a
two-predicate rule already has, and it breaks ties against predicates that appear
in many rules. What we observe: on a matched 600/200 split, CoupledSGA reaches
40.48\% R@20 against 44.78\% for SceneSayerODE, and in a controlled comparison,
enabling the penalty after two warm-up epochs lowers R@20 from 42.21\% to 35.18\%.
The second, which isolates the penalty, has the sign the analysis predicts. Under No Constraint, the
same change lowers the contradiction rate from 45.92\% to 4.07\%, and coupling
alone lowers it from 54.40\% to 45.92\%. On this measure, the penalty reduces contradictions at a recall cost, and With-Constraint recall reports only the cost.

Two next steps follow: a graph-level structural consistency score reported
alongside the contradiction rate, and the degree-normalised form of the penalty
suggested by Corollary~\ref{cor:degree}.

\begin{acks}
Suman Chowdhury and Rajesh P. Barnwal acknowledge partial financial support from
CSIR-CMERI, Durgapur, under Project MLP002613. Suman Chowdhury also acknowledges
financial support from the Council of Scientific and Industrial Research (CSIR),
India, through a Senior Research Fellowship (GATE), Grant
No.~GATE31/\allowbreak 0019(15628)/\allowbreak \mbox{2022-EMR-I}. AI-based tools
were used for copy-editing only; the authors take full responsibility for the
content.
\end{acks}

\bibliographystyle{ACM-Reference-Format}
\bibliography{references}


\begin{thebibliography}{24}


\ifx \showCODEN    \undefined \def \showCODEN     #1{\unskip}     \fi
\ifx \showISBNx    \undefined \def \showISBNx     #1{\unskip}     \fi
\ifx \showISBNxiii \undefined \def \showISBNxiii  #1{\unskip}     \fi
\ifx \showISSN     \undefined \def \showISSN      #1{\unskip}     \fi
\ifx \showLCCN     \undefined \def \showLCCN      #1{\unskip}     \fi
\ifx \shownote     \undefined \def \shownote      #1{#1}          \fi
\ifx \showarticletitle \undefined \def \showarticletitle #1{#1}   \fi
\ifx \showURL      \undefined \def \showURL       {\relax}        \fi
\providecommand\bibfield[2]{#2}
\providecommand\bibinfo[2]{#2}
\providecommand\natexlab[1]{#1}
\providecommand\showeprint[2][]{arXiv:#2}
\makeatletter
\@ifundefined{NAT@parse@date}{}{\let\NAT@parse@date@orig\NAT@parse@date}
\@ifundefined{NAT@parse@date}{}{\def\NAT@parse@date#1#2#3#4#5#6@@{\NAT@parse@date@orig#1#2#3#4#5#6@@\def\NAT@tempyear{0000}\def\NAT@tempexlab{{?}}\ifx\NAT@year\NAT@tempyear\ifx\NAT@exlab\NAT@tempexlab\def\NAT@date{[n.\,d.]}\else\edef\NAT@date{[n.\,d.]\NAT@exlab}\fi\fi}}
\makeatother

\bibitem[Alliegro et~al\mbox{.}(2025)]%
        {alliegro2025forescene}
\bibfield{author}{\bibinfo{person}{Antonio Alliegro},
  \bibinfo{person}{Francesca Pistilli}, \bibinfo{person}{Tatiana Tommasi},
  {and} \bibinfo{person}{Giuseppe Averta}.} \bibinfo{year}{2025}\natexlab{}.
\newblock \bibinfo{title}{{FORESCENE}: {FOREcasting} Human Activity via Latent
  {SCENE} Graphs Diffusion}.
\newblock
\showeprint[arxiv]{2503.06182}~[cs.CV]


\bibitem[Bengio et~al\mbox{.}(2009)]%
        {bengio2009curriculum}
\bibfield{author}{\bibinfo{person}{Yoshua Bengio},
  \bibinfo{person}{J{\'e}r{\^o}me Louradour}, \bibinfo{person}{Ronan
  Collobert}, {and} \bibinfo{person}{Jason Weston}.}
  \bibinfo{year}{2009}\natexlab{}.
\newblock \showarticletitle{Curriculum Learning}. In
  \bibinfo{booktitle}{\emph{Proceedings of the 26th Annual International
  Conference on Machine Learning (ICML)}}. \bibinfo{publisher}{ACM},
  \bibinfo{address}{New York, NY, USA}, \bibinfo{pages}{41--48}.
\newblock
\href{https://doi.org/10.1145/1553374.1553380}{doi:\nolinkurl{10.1145/1553374.1553380}}


\bibitem[Brody et~al\mbox{.}(2022)]%
        {brody2022gatv2}
\bibfield{author}{\bibinfo{person}{Shaked Brody}, \bibinfo{person}{Uri Alon},
  {and} \bibinfo{person}{Eran Yahav}.} \bibinfo{year}{2022}\natexlab{}.
\newblock \showarticletitle{How Attentive Are Graph Attention Networks?}. In
  \bibinfo{booktitle}{\emph{International Conference on Learning
  Representations (ICLR)}}. \bibinfo{publisher}{OpenReview.net}.
\newblock


\bibitem[Chen et~al\mbox{.}(2018)]%
        {chen2018neural}
\bibfield{author}{\bibinfo{person}{Ricky T.~Q. Chen}, \bibinfo{person}{Yulia
  Rubanova}, \bibinfo{person}{Jesse Bettencourt}, {and}
  \bibinfo{person}{David~K. Duvenaud}.} \bibinfo{year}{2018}\natexlab{}.
\newblock \showarticletitle{Neural Ordinary Differential Equations}. In
  \bibinfo{booktitle}{\emph{Advances in Neural Information Processing Systems
  (NeurIPS)}}, Vol.~\bibinfo{volume}{31}. \bibinfo{publisher}{Curran
  Associates, Inc.}, \bibinfo{address}{Red Hook, NY, USA},
  \bibinfo{pages}{6571--6583}.
\newblock


\bibitem[Chen et~al\mbox{.}(2019)]%
        {chen2019kern}
\bibfield{author}{\bibinfo{person}{Tianshui Chen}, \bibinfo{person}{Weihao Yu},
  \bibinfo{person}{Riquan Chen}, {and} \bibinfo{person}{Liang Lin}.}
  \bibinfo{year}{2019}\natexlab{}.
\newblock \showarticletitle{Knowledge-Embedded Routing Network for Scene Graph
  Generation}. In \bibinfo{booktitle}{\emph{Proceedings of the IEEE/CVF
  Conference on Computer Vision and Pattern Recognition (CVPR)}}.
  \bibinfo{publisher}{IEEE}, \bibinfo{address}{Piscataway, NJ, USA},
  \bibinfo{pages}{6163--6171}.
\newblock


\bibitem[Cong et~al\mbox{.}(2021)]%
        {cong2021sttran}
\bibfield{author}{\bibinfo{person}{Yuren Cong}, \bibinfo{person}{Wentong Liao},
  \bibinfo{person}{Hanno Ackermann}, \bibinfo{person}{Bodo Rosenhahn}, {and}
  \bibinfo{person}{Michael~Ying Yang}.} \bibinfo{year}{2021}\natexlab{}.
\newblock \showarticletitle{Spatial-Temporal Transformer for Dynamic Scene
  Graph Generation}. In \bibinfo{booktitle}{\emph{Proceedings of the IEEE/CVF
  International Conference on Computer Vision (ICCV)}}.
  \bibinfo{publisher}{IEEE}, \bibinfo{address}{Piscataway, NJ, USA},
  \bibinfo{pages}{16372--16382}.
\newblock


\bibitem[Feng et~al\mbox{.}(2023)]%
        {feng2023dsg}
\bibfield{author}{\bibinfo{person}{Shengyu Feng}, \bibinfo{person}{Hesham
  Mostafa}, \bibinfo{person}{Marcel Nassar}, \bibinfo{person}{Somdeb Majumdar},
  {and} \bibinfo{person}{Subarna Tripathi}.} \bibinfo{year}{2023}\natexlab{}.
\newblock \showarticletitle{Exploiting Long-Term Dependencies for Generating
  Dynamic Scene Graphs}. In \bibinfo{booktitle}{\emph{Proceedings of the
  IEEE/CVF Winter Conference on Applications of Computer Vision (WACV)}}.
  \bibinfo{publisher}{IEEE}, \bibinfo{address}{Piscataway, NJ, USA},
  \bibinfo{pages}{5130--5139}.
\newblock


\bibitem[Furnari and Farinella(2019)]%
        {furnari2019rolling}
\bibfield{author}{\bibinfo{person}{Antonino Furnari} {and}
  \bibinfo{person}{Giovanni~Maria Farinella}.} \bibinfo{year}{2019}\natexlab{}.
\newblock \showarticletitle{What Would You Expect? {A}nticipating Egocentric
  Actions with Rolling-Unrolling {LSTMs} and Modality Attention}. In
  \bibinfo{booktitle}{\emph{Proceedings of the IEEE/CVF International
  Conference on Computer Vision (ICCV)}}. \bibinfo{publisher}{IEEE},
  \bibinfo{address}{Piscataway, NJ, USA}, \bibinfo{pages}{6252--6261}.
\newblock


\bibitem[Gammulle et~al\mbox{.}(2019)]%
        {gammulle2019predicting}
\bibfield{author}{\bibinfo{person}{Harshala Gammulle}, \bibinfo{person}{Simon
  Denman}, \bibinfo{person}{Sridha Sridharan}, {and} \bibinfo{person}{Clinton
  Fookes}.} \bibinfo{year}{2019}\natexlab{}.
\newblock \showarticletitle{Predicting the Future: A Jointly Learnt Model for
  Action Anticipation}. In \bibinfo{booktitle}{\emph{Proceedings of the
  IEEE/CVF International Conference on Computer Vision (ICCV)}}.
  \bibinfo{publisher}{IEEE}, \bibinfo{address}{Piscataway, NJ, USA},
  \bibinfo{pages}{5562--5571}.
\newblock


\bibitem[Huang et~al\mbox{.}(2021)]%
        {huang2021cgode}
\bibfield{author}{\bibinfo{person}{Zijie Huang}, \bibinfo{person}{Yizhou Sun},
  {and} \bibinfo{person}{Wei Wang}.} \bibinfo{year}{2021}\natexlab{}.
\newblock \showarticletitle{Coupled Graph {ODE} for Learning Interacting System
  Dynamics}. In \bibinfo{booktitle}{\emph{Proceedings of the 27th ACM SIGKDD
  Conference on Knowledge Discovery and Data Mining (KDD)}}.
  \bibinfo{publisher}{ACM}, \bibinfo{address}{New York, NY, USA},
  \bibinfo{pages}{705--715}.
\newblock
\href{https://doi.org/10.1145/3447548.3467385}{doi:\nolinkurl{10.1145/3447548.3467385}}


\bibitem[Ji et~al\mbox{.}(2020)]%
        {ji2020actiongenome}
\bibfield{author}{\bibinfo{person}{Jingwei Ji}, \bibinfo{person}{Ranjay
  Krishna}, \bibinfo{person}{Li Fei-Fei}, {and} \bibinfo{person}{Juan~Carlos
  Niebles}.} \bibinfo{year}{2020}\natexlab{}.
\newblock \showarticletitle{Action Genome: Actions as Compositions of
  Spatio-Temporal Scene Graphs}. In \bibinfo{booktitle}{\emph{Proceedings of
  the IEEE/CVF Conference on Computer Vision and Pattern Recognition (CVPR)}}.
  \bibinfo{publisher}{IEEE}, \bibinfo{address}{Piscataway, NJ, USA},
  \bibinfo{pages}{10236--10247}.
\newblock


\bibitem[Loshchilov and Hutter(2019)]%
        {loshchilov2019decoupled}
\bibfield{author}{\bibinfo{person}{Ilya Loshchilov} {and}
  \bibinfo{person}{Frank Hutter}.} \bibinfo{year}{2019}\natexlab{}.
\newblock \showarticletitle{Decoupled Weight Decay Regularization}. In
  \bibinfo{booktitle}{\emph{International Conference on Learning
  Representations (ICLR)}}. \bibinfo{publisher}{OpenReview.net}.
\newblock


\bibitem[Manhaeve et~al\mbox{.}(2018)]%
        {manhaeve2018deepproblog}
\bibfield{author}{\bibinfo{person}{Robin Manhaeve}, \bibinfo{person}{Sebastijan
  Duman{\v{c}}i{\'c}}, \bibinfo{person}{Angelika Kimmig},
  \bibinfo{person}{Thomas Demeester}, {and} \bibinfo{person}{Luc De~Raedt}.}
  \bibinfo{year}{2018}\natexlab{}.
\newblock \showarticletitle{{DeepProbLog}: Neural Probabilistic Logic
  Programming}. In \bibinfo{booktitle}{\emph{Advances in Neural Information
  Processing Systems (NeurIPS)}}, Vol.~\bibinfo{volume}{31}.
  \bibinfo{publisher}{Curran Associates, Inc.}, \bibinfo{address}{Red Hook, NY,
  USA}, \bibinfo{pages}{3749--3759}.
\newblock


\bibitem[Nag et~al\mbox{.}(2023)]%
        {nag2023tempura}
\bibfield{author}{\bibinfo{person}{Sayak Nag}, \bibinfo{person}{Kyle Min},
  \bibinfo{person}{Subarna Tripathi}, {and} \bibinfo{person}{Amit~K.
  Roy-Chowdhury}.} \bibinfo{year}{2023}\natexlab{}.
\newblock \showarticletitle{Unbiased Scene Graph Generation in Videos}. In
  \bibinfo{booktitle}{\emph{Proceedings of the IEEE/CVF Conference on Computer
  Vision and Pattern Recognition (CVPR)}}. \bibinfo{publisher}{IEEE},
  \bibinfo{address}{Piscataway, NJ, USA}, \bibinfo{pages}{22803--22813}.
\newblock


\bibitem[Peddi et~al\mbox{.}(2025)]%
        {peddi2025impartail}
\bibfield{author}{\bibinfo{person}{Rohith Peddi}, \bibinfo{person}{Saurabh},
  \bibinfo{person}{Ayush~Abhay Shrivastava}, \bibinfo{person}{Parag Singla},
  {and} \bibinfo{person}{Vibhav Gogate}.} \bibinfo{year}{2025}\natexlab{}.
\newblock \showarticletitle{Towards Unbiased and Robust Spatio-Temporal Scene
  Graph Generation and Anticipation}. In \bibinfo{booktitle}{\emph{Proceedings
  of the IEEE/CVF Conference on Computer Vision and Pattern Recognition
  (CVPR)}}. \bibinfo{publisher}{IEEE}, \bibinfo{address}{Piscataway, NJ, USA},
  \bibinfo{pages}{8648--8657}.
\newblock


\bibitem[Peddi et~al\mbox{.}(2024)]%
        {peddi2024scenesayer}
\bibfield{author}{\bibinfo{person}{Rohith Peddi}, \bibinfo{person}{Saksham
  Singh}, \bibinfo{person}{Saurabh}, \bibinfo{person}{Parag Singla}, {and}
  \bibinfo{person}{Vibhav Gogate}.} \bibinfo{year}{2024}\natexlab{}.
\newblock \showarticletitle{Towards Scene Graph Anticipation}. In
  \bibinfo{booktitle}{\emph{Computer Vision -- ECCV 2024}}
  \emph{(\bibinfo{series}{Lecture Notes in Computer Science},
  Vol.~\bibinfo{volume}{15146})}. \bibinfo{publisher}{Springer},
  \bibinfo{address}{Cham}, \bibinfo{pages}{159--175}.
\newblock
\href{https://doi.org/10.1007/978-3-031-73223-2_10}{doi:\nolinkurl{10.1007/978-3-031-73223-2_10}}


\bibitem[Poli et~al\mbox{.}(2019)]%
        {poli2019graph}
\bibfield{author}{\bibinfo{person}{Michael Poli}, \bibinfo{person}{Stefano
  Massaroli}, \bibinfo{person}{Junyoung Park}, \bibinfo{person}{Atsushi
  Yamashita}, \bibinfo{person}{Hajime Asama}, {and} \bibinfo{person}{Jinkyoo
  Park}.} \bibinfo{year}{2019}\natexlab{}.
\newblock \bibinfo{title}{Graph Neural Ordinary Differential Equations}.
\newblock
\showeprint[arxiv]{1911.07532}~[cs.LG]


\bibitem[Rusch et~al\mbox{.}(2022)]%
        {rusch2022graph}
\bibfield{author}{\bibinfo{person}{T.~Konstantin Rusch}, \bibinfo{person}{Ben
  Chamberlain}, \bibinfo{person}{James Rowbottom}, \bibinfo{person}{Siddhartha
  Mishra}, {and} \bibinfo{person}{Michael Bronstein}.}
  \bibinfo{year}{2022}\natexlab{}.
\newblock \showarticletitle{Graph-Coupled Oscillator Networks}. In
  \bibinfo{booktitle}{\emph{Proceedings of the 39th International Conference on
  Machine Learning (ICML)}} \emph{(\bibinfo{series}{Proceedings of Machine
  Learning Research}, Vol.~\bibinfo{volume}{162})}. \bibinfo{publisher}{PMLR},
  \bibinfo{pages}{18888--18909}.
\newblock


\bibitem[Sigurdsson et~al\mbox{.}(2016)]%
        {sigurdsson2016hollywood}
\bibfield{author}{\bibinfo{person}{Gunnar~A. Sigurdsson},
  \bibinfo{person}{G{\"u}l Varol}, \bibinfo{person}{Xiaolong Wang},
  \bibinfo{person}{Ali Farhadi}, \bibinfo{person}{Ivan Laptev}, {and}
  \bibinfo{person}{Abhinav Gupta}.} \bibinfo{year}{2016}\natexlab{}.
\newblock \showarticletitle{Hollywood in Homes: Crowdsourcing Data Collection
  for Activity Understanding}. In \bibinfo{booktitle}{\emph{Computer Vision --
  ECCV 2016}} \emph{(\bibinfo{series}{Lecture Notes in Computer Science},
  Vol.~\bibinfo{volume}{9905})}. \bibinfo{publisher}{Springer},
  \bibinfo{address}{Cham}, \bibinfo{pages}{510--526}.
\newblock
\href{https://doi.org/10.1007/978-3-319-46448-0_31}{doi:\nolinkurl{10.1007/978-3-319-46448-0_31}}


\bibitem[Suhail et~al\mbox{.}(2021)]%
        {suhail2021energy}
\bibfield{author}{\bibinfo{person}{Mohammed Suhail}, \bibinfo{person}{Abhay
  Mittal}, \bibinfo{person}{Behjat Siddiquie}, \bibinfo{person}{Chris
  Broaddus}, \bibinfo{person}{Jayan Eledath}, \bibinfo{person}{Gerard Medioni},
  {and} \bibinfo{person}{Leonid Sigal}.} \bibinfo{year}{2021}\natexlab{}.
\newblock \showarticletitle{Energy-Based Learning for Scene Graph Generation}.
  In \bibinfo{booktitle}{\emph{Proceedings of the IEEE/CVF Conference on
  Computer Vision and Pattern Recognition (CVPR)}}. \bibinfo{publisher}{IEEE},
  \bibinfo{address}{Piscataway, NJ, USA}, \bibinfo{pages}{13936--13945}.
\newblock


\bibitem[Xu et~al\mbox{.}(2017)]%
        {xu2017scene}
\bibfield{author}{\bibinfo{person}{Danfei Xu}, \bibinfo{person}{Yuke Zhu},
  \bibinfo{person}{Christopher~B. Choy}, {and} \bibinfo{person}{Li Fei-Fei}.}
  \bibinfo{year}{2017}\natexlab{}.
\newblock \showarticletitle{Scene Graph Generation by Iterative Message
  Passing}. In \bibinfo{booktitle}{\emph{Proceedings of the IEEE Conference on
  Computer Vision and Pattern Recognition (CVPR)}}. \bibinfo{publisher}{IEEE},
  \bibinfo{address}{Piscataway, NJ, USA}, \bibinfo{pages}{5410--5419}.
\newblock


\bibitem[Yang et~al\mbox{.}(2025)]%
        {yang2025fdsg}
\bibfield{author}{\bibinfo{person}{Yi Yang}, \bibinfo{person}{Yuren Cong},
  \bibinfo{person}{Hao Cheng}, \bibinfo{person}{Bodo Rosenhahn}, {and}
  \bibinfo{person}{Michael~Ying Yang}.} \bibinfo{year}{2025}\natexlab{}.
\newblock \bibinfo{title}{{FDSG}: Forecasting Dynamic Scene Graphs}.
\newblock
\showeprint[arxiv]{2506.01487}~[cs.CV]


\bibitem[Zareian et~al\mbox{.}(2020)]%
        {zareian2020bridging}
\bibfield{author}{\bibinfo{person}{Alireza Zareian}, \bibinfo{person}{Svebor
  Karaman}, {and} \bibinfo{person}{Shih-Fu Chang}.}
  \bibinfo{year}{2020}\natexlab{}.
\newblock \showarticletitle{Bridging Knowledge Graphs to Generate Scene
  Graphs}. In \bibinfo{booktitle}{\emph{Computer Vision -- ECCV 2020}}.
  \bibinfo{publisher}{Springer}, \bibinfo{address}{Cham},
  \bibinfo{pages}{606--623}.
\newblock
\href{https://doi.org/10.1007/978-3-030-58592-1_36}{doi:\nolinkurl{10.1007/978-3-030-58592-1_36}}


\bibitem[Zellers et~al\mbox{.}(2018)]%
        {zellers2018neural}
\bibfield{author}{\bibinfo{person}{Rowan Zellers}, \bibinfo{person}{Mark
  Yatskar}, \bibinfo{person}{Sam Thomson}, {and} \bibinfo{person}{Yejin Choi}.}
  \bibinfo{year}{2018}\natexlab{}.
\newblock \showarticletitle{Neural Motifs: Scene Graph Parsing with Global
  Context}. In \bibinfo{booktitle}{\emph{Proceedings of the IEEE Conference on
  Computer Vision and Pattern Recognition (CVPR)}}. \bibinfo{publisher}{IEEE},
  \bibinfo{address}{Piscataway, NJ, USA}, \bibinfo{pages}{5831--5840}.
\newblock


\end{thebibliography}

\end{document}